# Modeling uncertain and vague knowledge in possibility and evidence theories

Didier DUBOIS - Henri PRADE

L.S.I., Université Paul Sabatier, 118 route de Narbonne 31062 TOULOUSE Cédex (FRANCE)
Tel. : 61.55.69.42

**Abstract** : This paper advocates the usefulness of new theories of uncertainty for the purpose of modeling some facets of uncertain knowledge, especially vagueness, in A.I.. It can be viewed as a partial reply to Cheeseman's (among others) defense of probability.

## 0. Introduction

In spite of the growing bulk of works dealing with deviant models of uncertainty in artificial intelligence (e.g. [14]), there is a strong reaction of classical probability tenants ([1]-[2] and [16] for instance), claiming that new uncertainty theories are "at best unnecessary, and at worst misleading" [1]. Interestingly enough, however, the trend to go beyond probabilistic models of subjective uncertainty is emerging even in the orthodox field of decision theory in order to account for systematic deviations of human behavior from the expected utility models. This paper tries to reconcile the points of view of probability theory and those of two presently popular alternative settings : possibility theory [22], [5] and the theory of evidence [19]. The focus is precisely on the representation of subjective uncertain knowledge. We try to explain why probability measures cannot account for all facets of uncertainty, especially partial ignorance, imprecision, vagueness, and how the other theories can do the job, without rejecting the laws of probability when they apply.

## 1. Representing uncertainty

Let $\mathcal{P}$ be a Boolean algebra of propositions, denoted by a, b, c... We assume that an uncertain piece of information can be represented by means of a logical proposition to which a number, conveying the level of uncertainty, is attached. The meaning of this number is partly a matter of context. Following Cheeseman [1], g(a) is here supposed to reflect an "entity's opinion about the truth of a, given the available evidence". Let g(a) denote the number attached to a. g is supposed to range over [0,1], and satisfies limit conditions g(0) = 0 and g(1) = 1, where **0** and **1** denote the contradiction and the tautology respectively.

### 1.1 Some limitations of a probabilistic model of subjective uncertainty

If g is assumed to be a probability measure, then g(a) = 1 means that a is certainly true, while g(a) = 0 means that a is certainly false. This convention does not allow for the modeling of partial ignorance in a systematic way. Especially it may create some problems in the presence of imprecise evidence, when part of it neither supports nor denies a. First, to allocate an amount of probability P(a) to a proposition a compels you to allocate 1-P(a) to the converse proposition 'not a' (¬a). It implies that P(a)=1 means "a is true" and P(a)=0 means "a is false". In case of total ignorance about a, you are bound to let P(a) = 1-P(¬a)=.5. If you are equally ignorant about the truth of another proposition b then you must do it again : P(b)=1 - P(¬b) = .5. Now assuming that a logically entails b (¬a ∨ b is the tautology **1**), you come up with the result that P(¬a ∧ b)= P(b) - P(a)=0 ! This is a paradox : how can your ignorance allow you to discard the possibility that ¬a ∧ b be true ! For instance a="horse n° 4 will win the race" b="horse with an even number wins the race", and the probabilistic modelling of ignorance leads you to the surprising statement "no horse with an even number other than 4 will win the race". What is pointed out is that probability theory

81

offers no stable, consistent reference value for the modelling of ignorance. This issue is important since even experts are often in situation of partial ignorance (e.g. on the precise value of a probability !). The way a question is answered depends upon the existence of further questions which reveal more about the structure of the set of possible answers. This is not psychologically satisfying.

Of course probability theory tenants would argue that this is not the proper way of applying probability theory. First you should make sure about how many horses are in the race and then your ignorance leads you to uniformly distributing your probabilities among the horses following the principle of maximum entropy [13]. This is good for games of chance, when it is known that dice have only 6 facets. But in the expert system area, they usually play with objects without exactly knowing their number of facets. An important part of probabilistic modelling consists of making up a set of exhaustive, mutually exclusive alternatives *before* assessing probabilities. For instance, in the horse example, this set must account for the constraint "a implies b"; it leaves only 3 alternatives: $a$, $\neg a \wedge b$, $\neg b$ to which probabilities must be assigned. The difficulty in A.I. is that this job is generally impossible to do, at least to do once for all. Knowledge bases must allow for evolving representations of the universe, that is the discovery of new unexpected events, without having to reconstruct a new knowledge base out of nothing, upon such occurrences. At a given point in the construction of a knowledge base, the expert may not be aware of all existing alternatives. Of course you can always use a special dummy alternative such as "all other possibilities". But the uniformly distributed probability assignment derived from Laplace's insufficient reason principle is not very convincing in that case ! Moreover the expert may not always be able to precisely describe the alternatives he is aware of, so that these descriptions may overlap each other and the mutual exclusiveness property needed in probability theory, is lost. Despite this not very nice setting, one must still be able to perform uncertain reasoning in knowledge bases ! Quoting Cheeseman [1], "if the problem is undefined probability theory cannot say something useful". But artificial intelligence tries to address partially undefined problems which human experts can cope with. This presence of imprecision forces us out of the strict probabilistic setting - this does not mean rejecting it, but enlarging it.

Even when an exhaustive set of mutually exclusive alternatives is available it is questionable to represent the state of total or partial ignorance by means of an uniformly distributed probability measure. The most recent justification for this latter approach seems to be the principle of maximum entropy (e.g. [13]). However, in the mind of some entropy principle tenants, there seems to be a confusion between two problems : one of representing a state of knowledge and one of making a decision. The kind of information measured by Shannon's entropy is the amount of <u>uncertainty</u> (pervading a probability assignment) regarding which outcome is likely to occur next, or what is the best decision to be made on such ground. The word "information" refers here to the existence of reasons to choose one alternative <u>against</u> another. Especially you are equally uncertain about what will happen when flipping a coin whether you totally ignore the mechanical properties of this particular coin, or you have made 1,000,000 experiments with the coin, and it has led to an equal amount of heads and tails. You are in the same state of uncertainty but certainly not in the same state of knowledge : in the case of total ignorance, you have <u>no</u> evidence about the coin ; in the second situation, you are not far from having the <u>maximal</u> amount of evidence, although being uncertain due to randomness. In artificial intelligence problems we want to have a model of our state of knowledge in which ignorance is carefully distinguished from randomness. This is important because when uncertainty is due to ignorance, there is some hope of improving the situation by getting more evidence (flip the coin, for instance). When uncertainty is due to experimentally observed randomness, it seems difficult to remove it by getting more information. Distinguishing between ignorance and randomness may be useful for the purpose of devising knowledge based systems equipped with self-explanation capabilities. The maximal entropy principle was not invented to take care of this distinction, and looks irrelevant as far as the representation of knowledge is concerned. It may prove more useful for decision support-systems than for approximate reasoning systems.

### 1.2 Belief functions

In the situation of partial ignorance the probability of **a** is only imprecisely known, and can be expressed as an interval range [C(**a**), Pl(**a**)] whose lower bound can be viewed as a degree of certainty (or belief) of **a**, while the upper bound represents a grade of plausibility (or possibility) of **a**, i.e. the extent to which **a** cannot be denied. Total ignorance about a is observed when there is a total lack of certainty (C(**a**) = 0) and complete possibility (Pl(**a**)=1) for **a**. A natural assumption is to admit that the evidence which supports a also denies 'not a' (¬**a**). This modeling assumption leads to the convention

$$C(\mathbf{a}) = 1 - Pl(\neg \mathbf{a}) \quad (1)$$

which is in agreement with g(a) = 1 - g(¬a), where g(a) and g(¬a) are imprecisely known probabilities. This equality also means that the certainty of **a** is equivalent to the impossibility of ¬**a**. The framework of probability theory does not allow for modelling the difference between possibility and certainty, as expressed by (1). Functions C and Pl are usually called lower and upper probabilities, when considered as bounds on an unknown probability measure. See [20] and [8] for surveys on lower and upper probabilities.

Using (1), the knowledge of the certainty function C over the Boolean algebra of propositions $\mathcal{P}$ is enough to reconstruct the plausibility function Pl. Especially the amount of uncertainty pervading **a** is summarized by the two numbers C(**a**) and C(¬**a**). They are such that C(**a**) + C(¬**a**) ≤ 1, due to (1). The above discussion leads to the following conventions, for interpreting the number C(**a**) attached to **a** :

i) C(**a**)=1 means that **a** is certainly true ; ii) C(¬**a**)=1 ⇔ Pl(**a**)=0 means that **a** is certainly false.
iii) C(**a**)=C(¬**a**) = 0 (i.e. Pl(**a**)=1) means total ignorance about a. In other words a is neither supported nor denied by any piece of available evidence. This is a self-consistent, absolute reference point for expressing ignorance.
iv) C(**a**)=C(¬**a**)=0.5 (i.e. Pl(**a**)=0.5) means maximal probabilistic uncertainty about a. In other words the available evidence can be shared in two equal parts : one which supports a and the other which denies it. This is the case of pure randomness in the occurence of a.

Note that total ignorance implies that we are equally uncertain about the truth of a and ¬a, as well as when C(**a**)=C(¬**a**)=.5. In other words ignorance implies uncertainty about the truth of **a**, but the converse is not true. Namely, in the probabilistic case, we have a lot of information, but we are still completely uncertain. Total uncertainty is more generally observed whenever C(**a**)=C(¬**a**)∈[0,0.5]; then the amount of ignorance is assessed by 1-2 C(**a**).

The mathematical properties of C depend upon the way the available evidence is modelled and related to the certainty function. In Shafer theory, a body of evidence ($\mathcal{F}$,m) is composed of a subset $\mathcal{F} \subseteq \mathcal{P}$ of n focal propositions, each being attached a relative weight of confidence m($a_i$). For all $a_i \in \mathcal{F}$, m($a_i$) is a positive number in the unit interval, and it holds

$$\sum_{i=1,n} m(a_i) = 1 \quad (2) \quad ; \quad m(0) = 0 \quad (3)$$

(3) expresses the fact that no confidence is committed to the contradictory proposition. The weight m(**1**) possibly granted to the tautology represents the amount of total ignorance since the tautology does not support nor deny any other proposition. The fact that a proposition **a** supports another proposition **b** is formally expressed by the logical entailment, i.e. **a** → **b** ( = ¬**a** ∨ **b** ) = 1. Let S(a) be the set of propositions supporting a other than the contradiction **0**. The function C(a) is called a *belief function* in the sense of Shafer (and denoted 'Bel') if and only if there is a body of evidence ($\mathcal{F}$,m) such that

$$\forall a,\ Bel(a) = \sum_{a_i \in S(a)} m(a_i) \quad (4) \quad ; \quad \forall a,\ Pl(a) = \sum_{a_i \in S(\neg a)^c - \{0\}} m(a_i) \quad (5)$$

where 'c' denotes complementation. Clearly, when the focal elements are only atoms of the Boolean algebra $\mathcal{P}$ (i.e. $S(a_i)=\{a_i\}$, for all i=1,n) then $\forall a$, $S(\neg a)=S(a)^c-\{0\}$, and $Pl(a)=Bel(a),\forall a$. We recover a probability measure on $\mathcal{P}$. In the general case the quantity $Pl(a) - Bel(a)$ represents the amount of imprecision about the probability of a. Interpreting the Boolean algebra $\mathcal{P}$ as the set of subsets of a referential set $\Omega$, the atoms of $\mathcal{P}$ can be viewed as the singletons of $\Omega$ and interpreted as possible worlds, one of which is the actual world. Then a focal element $a_i$ whose model is the subset $M(a_i) = A_i \subseteq \Omega$ corresponds to the statement : the actual world is in $A_i$ with probability $m(A_i)$. When $A_i$ is not a singleton, this piece of information is said to be imprecise, because the actual world can be anywhere within $A_i$. When $A_i$ is a singleton, the piece of information is said to be precise. Clearly, Bel = Pl is a probability measure if and only if the available evidence is precise (but generally scattered between several disjoint focal elements viewed as singletons).

Note that although Bel(a) and Pl(a) are respectively lower and upper probabilities, the converse is not true, that is any interval-valued probability cannot be interpreted as a pair of belief and plausibility functions in the sense of (4) and (5) (see [20], [8]).

### 1.3. Possibility measures

When two propositions **a** and **b** are such that $a \in S(b)$, we write $a \vdash b$, and $\vdash$ is called the entailment relation. Note that $\vdash$ is reflexive and transitive and equips the set $\mathcal{F}$ of focal elements with a partial ordering structure. When $\mathcal{F}$ is linearly ordered by $\vdash$, i.e., $\mathcal{F} =\{a_1,...,a_n\}$ where $a_i \vdash a_{i+1}$, i = 1,n-1, the belief and plausibility functions Bel and Pl satisfy [21] :

$$\forall a, \forall b,\ Bel(a \wedge b) = \min(Bel(a), Bel(b)) \quad ; \quad Pl(a \vee b) = \max(Pl(a), Pl(b)) \quad (6)$$

Formally, the plausibility function is a possibility measure in the sense of Zadeh [22]. Then

$$\max(Pl(a), Pl(\neg a)) = 1 \quad (7) \quad ; \quad \min(Bel(a), Bel(\neg a)) = 0 \quad (8) \quad ; \quad Bel(a) > 0 \Rightarrow Pl(a) = 1 \quad (9)$$

In the following possibility measures are denoted $\Pi$ for the sake of clarity. The dual measure through (1) is then denoted N and called a necessity measure [5]. Zadeh [22] introduces possibility measures from so-called possibility distributions, which are mappings from $\Omega$ to [0,1], denoted $\pi$. A possibility and the dual necessity measure are then obtained as

$$\forall A \subseteq \Omega,\ \Pi(A) = \sup\{\pi(w) \mid w \in A\} \quad (10) \quad ; \quad N(A) = \inf\{1 - \pi(w) \mid w \in A^c\} \quad (11)$$

and we then have $\pi(w) = \Pi(\{w\}), \forall w$. The function $\pi$ can be viewed as a generalized characteristic function, i.e. the membership function of a fuzzy set F. Let $F_\alpha$ be the $\alpha$-cut of F. i.e., the subset $\{w \mid \mu_F(w) \geq \alpha\}$ with $\pi = \mu_F$. It is easy to check that in the finite case the set of $\alpha$-cuts $\{F_\alpha \mid \alpha \in (0,1]\}$ is the set $\mathcal{F}$ of focal elements of the possibility measure $\Pi$. Moreover, let $\pi_1 = 1 > \pi_2 ... > \pi_m$ be the set of distinct values of $\pi(w)$, let $\pi_{m+1} = 0$ by convention, and $A_i$ be the $\pi_i$-cut of F, i = 1,m. Then the basic probability assignment m underlying $\Pi$ is completely defined in terms of the possibility distribution $\pi$ as [4] :

$$m(A_i) = \pi_i - \pi_{i+1}\ i = 1,m \quad ; \quad m(A) = 0\ \text{otherwise} \quad (12)$$

N.B. : Interpreting N(a) as a belief degree as in MYCIN and N($\neg$a) as a degree of disbelief in a, (6), (7) and (8) are assumed in MYCIN to be valid. Hence, as pointed out in [5], MYCIN's treatment of uncertainty is partly consistent with possibility theory.

Among the unjustified criticisms of possibility theory is the statement by Cheeseman [2] that it "contains rules such as $\Pi(a \wedge b) = \min(\Pi(a), \Pi(b))$". This is wrong. Only an inequality, $\Pi(a \wedge b) \leq \min(\Pi(a), \Pi(b))$ is valid, generally, just as in probability theory. The above equality holds in very special cases [5], namely when **a** and **b** pertain to variables which do not interact with each other. Possibility logic relies on (6) but certainly not on the equality mentioned by Cheeseman [2]. Possibility logic, just as probabilistic logic is not truth-functional. It can even be proved that in the presence of uncertainty, the two equalities $\Pi(a \vee b) = \max(\Pi(a), \Pi(b))$ and $\Pi(a \wedge b) = \min(\Pi(a), \Pi(b))$ ∀a ∀b, are inconsistent. So it is wrong to claim, as Cheeseman [2] does, that possibility theory "assumes a state of maximal dependence between components".

The conventions adopted in this paper to quantify uncertainty can be visually illustrated by a rectangle triangle in a Cartesian coordinate system that is reminiscent of Rollinger [18]'s conventions (see Figure 1). Any point in the triangle has coordinates (Bel(a), Bel(¬a)).

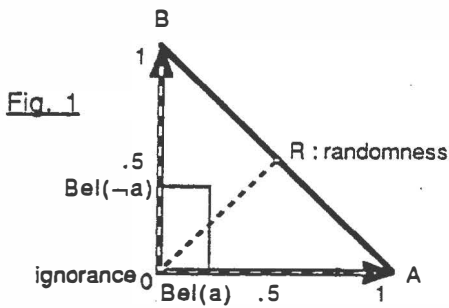

Fig. 1

The x-axis (resp. y-axis) quantifies support for **a** (resp. ¬**a**). Hence any point in the triangle AOB expresses a state of knowledge, viewed as a convex combination of total certainty for **a** (vertex A), for ¬**a** (vertex B), and total ignorance (vertex O). Probabilistic knowledge lies on the A-B segment, and possibilistic knowledge lies on the coordinate axes. Totally uncertain states are located on the segment bounded by O and R, R being the mid-point of the probabilistic segment.

## 2. A short discussion of Cox's axiomatic framework for probability

Traditionally, a degree of probability can be interpreted as
- either the ratio between the number of outcomes which realize an event over the number of possible outcomes. This is good for games of chance.
- or the frequency of occurrence of an event, after a sufficient (theoretically infinite) number of trials. This is the frequentist view.
- or a numerical translation of an entity's opinion about the truth of a proposition, given the available evidence as termed in [2]. This is the subjectivist view, which has been expressed in various settings : axiomatic [3], pragmatic (this is the Bayesian approach to betting behavior, and qualitative comparative probability (see [11]) ; we shall focus only on Cox's axiomatic view and ask whether such a view exist for belief and possibility functions ; see [10] for details. See [9] for other views of these functions.

Cox [3] tried to prove that probability measures were the only possible numerical model of "reasonable expectation". He started from the following requirements. Letting f(**b** | **a**) be a measure of the reasonable credibility of the proposition **b** when the proposition **a** is known to be true. Cox proposes two basic axioms :

C1. there is some operation • such that $f(c \wedge b | a) = f(c | b \wedge a) \cdot f(b | a)$

C2. there is a function S such that $f(\neg b | a) = S(f(b | a))$

The following additional technical requirements are used in [3] :

C3. • and S have both continuous second order derivatives

Then, f is proved to be isomorphic to a probability measure. The purely technical assumption (C3) is very strong and cannot be justified on common sense arguments. For instance • = minimum is a solution of C1 which does not violate the algebra of propositions, but certainly violates C3. In fact the unicity result does not require C3, which can be relaxed into a more intuitive continuity and monotonicity assumption. Cox's unicity result <u>only</u> holds under <u>strict</u> monotonicity assumption [10]. Cheeseman [2] proposes Cox's results as a formal proof that only probability measures are reasonable for the modeling of subjective uncertainty. His claim can be disputed. Although C1 sounds very sensible as a definition of conditional credibility function, C2 explicitly states that only one number is enough to describe both the uncertainties of **b** and ¬**b**. Clearly, this statement rules out the ability to distinguish between the notions of possibility and certainty. This distinction is the very purpose of belief functions, possibility measures, and any kind of upper and lower probability system. Hence the unicity result is not so surprising, and Cox's setting, although being an interesting attempt at recovering probability measures from a purely non frequentist argument does not provide the ultimate answer to uncertainty modelling problems.

85

## 3. Modeling vagueness

Cheeseman [2] has proposed a nice definition of vagueness which we shall adopt, namely: "vagueness is uncertainty about meaning and can be represented by a probability distribution over possible meanings". However, contrastedly with [2], we explain why this view is completely consistent with fuzzy sets and with the set-theoretic view of belief functions.

### 3.1 Membership functions and intermediary grades of truth

Consider the statement "Mary is young", represented by the logical formula $a$ = young($x$) where $x$ stands for the actual value of Mary's age. The set of possible worlds is an age scale $\Omega$. A rough model of "young" consists in letting $M(a) = I_c$, an interval contained in $\Omega$, for instance the interval [0,25] years. $I_c$ is called the meaning of "young" the subscript c indicates the context where the information arises. Then "Mary is young" is true if $x \in I_c$ and false otherwise. Vagueness arises when there is uncertainty regarding which interval in $\Omega$ properly translates "young". Let $m(A)$ be the probability that $I_c = A$. $m(A)$ can be a subjective probability obtained by asking a single individual, or can be a frequency if it reflects a proportion of individuals thinking that A properly expresses "young".

Knowing the age $x$ of Mary, the grade of truth of the statement "Mary is young" is defined by the grade of membership $\mu_{young}(x)$ defined

$$\mu_{young}(x) = \sum_{A \,:\, x \in A} m(A) \qquad (13)$$

$\mu_{young}(x)$ estimates the extent to which the value $x$ is compatible with the meaning of "young", formally expressed under the form of a random <u>set</u> i.e. a body of evidence in the sense of Shafer [15], [6]. This view is a translation of Cheeseman [2]'s definition of vagueness. It becomes exactly Zadeh's definition of a fuzzy set as soon as the family $\{A_i \mid m(A_i) > 0\}$ is a nested family so that the knowledge of the membership function $\mu_{young}$ is equivalent to that of the probabilities $m(A_i)$, because (12) is equivalent to (13) (see [6]). Of course this nested property is not always completely valid in practice, especially if the $A_i$'s come from different individuals. However consonant (nested) approximations of dissonant bodies of evidence exist [6], which are especially very good when $\cap_i A_i \neq \emptyset$, a usually satisfied consistency property which expresses that there is an age in $\Omega$, totally compatible with "young". Hence a fuzzy set, with membership function $\mu : \Omega \rightarrow [0,1]$, can always be used as an approximation of a random set. However the word "random" may be inappropriate when the $m(A_i)$'s do not express frequencies. One might prefer the term "convex combination of sets", which is more neutral but lengthy.

Cheeseman [2] claims that a membership function $\mu_{young}$ is nothing but a conditional probability P(young | $x$). This claim is not very well founded. Indeed the existence of a quantity such as P(young | $x$) underlies the assumption that 'young' represents an event in the usual sense of probability. Hence 'young' is a given subset $I_c$ of $\Omega$. But then the value of P(young | $x$) is either 0 ($x \notin I_c$) or 1 ($x \in I_c$). This is because $x$ and $I_c$ belong to the same universe $\Omega$. Admitting that P(young | $x$) $\in$ (0,1) leads to admit that 'young' is not a standard event but has a membership function; Zadeh [21] has introduced the notion of the probability of a fuzzy event, defined, in the tradition of probability theory, as the expectation of the membership function $\mu_{young}$.

$$P(young) = \int_\Omega \mu_{young}(u) \, dP(u) \qquad (14)$$

This definition assumes that the a priori knowledge on Mary's age $u$ is given by a probability measure P and is exactly equation (5) of Cheeseman [2] p. 95. When calculating P(young | $x$), Mary's age is known ($u = x$) and P(young) = P(young | $x$) = $\mu_{young}(x)$. But because in that case the available information ($u = x$) is deterministic, P(. | $x$) is a Dirac measure (P($u|x$) = 1 if

86

u=x, and 0 otherwise) and writing P(young | x) is just a matter of convention, one could write Bel(young | x), $\Pi$(young | x) etc... as well, since a Dirac measure is also a particular case of belief function, possibility measure, etc... ! If one admits as Hisdal [12] that concepts like 'young' have clear-cut meanings for single individuals, but become fuzzy when a group of individuals is considered, then P(young|x) reflects the proportion of individuals that claim that x agrees with their notion of young; then $\mu_{young}$ can be given a probabilistic interpretation (as a likelihood function). This is also in accordance with (13) if m(A) denotes the proportion of individuals for whom 'young' means A.

### 3.2 Uncertainty and graded truth

Note that $\mu_{young}(x)$ is really a grade of truth of the statement "Mary is young" knowing that x is the actual age of Mary. It is not a grade of uncertainty about the truth of the statement. The statement is <u>partially true</u> ($\mu_{young}(x) \in (0,1)$) as long as x is a borderline age for the concept young (e.g. x = 30 in a given context). We can get grades of uncertainty about truth when reversing the problem, namely finding whether Mary's age (a variable u) is x, given that "Mary is young". The statement "Mary's age is x" can only be true or false but its truth may not be known. Let g(young,x) be the grade of possibility (or upper probability) that Mary is young <u>and</u> her age is x. We now wish to compute g(x|young) i.e. the degree of possibility that Mary's age is x <u>given that</u> she is young. Using Cox's axiom C1 we get

$$g(young, x) = g(young | x) * g(x) = g(x | young) * g(young) \quad (15)$$

where $*$ is some unspecified (non-decreasing, associative) combination operation such that $a * 1 = a$, $0 * 0 = 0$. This decomposition is the basis of conditional probability definition, with g = P, $*$ = product. It is by definition valid for any measure of uncertainty, consistently with Cox [3]'s proposal. Let us evaluate various terms in order to calculate g(x | young).

g(young | x) = $\mu_{young}(x)$ as shown above, regardless of the nature of g.

g(A) expresses the a priori knowledge about Mary's age belonging to A. We shall assume the state of total ignorance, so that g(A) = 1 $\forall A \neq \emptyset$, since g(A) is a degree of plausibility, consistently with section 1. Particularly g(x) = 1 and g(young) = 1.

As a consequence (15) leads to a well known equation in likelihood theory :

$$g(x | young) = g(young | x) = \mu_{young}(x) \quad (16)$$

When g is, stricto sensu, a possibility measure, (16) is exactly Zadeh's basic identity [22] which translates a verbal statement (such as "Mary is young") into a possibility distribution $\pi$ restricting the age u of Mary, defined by $\quad \pi(u = x) = \mu_{young}(x) \quad (17)$
i.e. $\mu_{young}(x)$ is interpreted as the grade of <u>possibility</u> that u = x given that "Mary is young" ; it is <u>not</u> the degree of truth of 'u = x'. Contrastedly, the degree of certainty C(x | young) can be computed using (1) and Dempster-Shafer framework as

$$C(x | young) = \sum_{A \subseteq \{x\}} m(A) = m(\{x\}) \quad (18)$$

i.e. C(x | young) = 0 (no certainty) generally because "young" usually refers to an age interval, and not to a precise age, so that m seldom bears on singletons. Hence, contrary to what is claimed in [12], conditional distributions of x|$\lambda$ and $\lambda$|x (where $\lambda$ is a fuzzy predicate) are distinct in possibility theory, provided that we consider both measures of possibility and necessity.

So, Cheeseman [2]'s view of vagueness looks consistent with possibility theory, much more than with probability theory when the a priori knowledge about the variable underlying a vague predicate is not available. Note that if this a priori knowledge were indeed available under the form of a probability measure P then (15) is specialized into

$$P(x | young) = \mu_{young}(x) \cdot [P(x) / P(young)] \quad (19)$$

which indeed defines a probability measure on $\Omega$ that is a revision of the a priori probability

87

P on the basis of the fuzzy event "Mary is young". If this vague statement were precise, i.e. "Mary's age is in $I_c$" then (19) becomes the well known conditioning formula

$$P(x|I_c) = P(x) / P(I_c) \text{ if } x \in I_c \quad ; \quad P(x|I_c) = 0 \text{ otherwise} \quad (20)$$

so (20), (19) and (17) can be regarded as particular consequences of Cox's axiom C1.

### 3.3 Uncertain evidence

Suppose as in Cheeseman [2] that the following information about Mary is that "Mary is probably young". This statement is denoted S. Cheeseman [2] models this statement by means of a conditional probability P(young | S) where once more the events appearing in the probability are not very well defined. Surprizingly, no attempt is made to apply the maximum entropy principle to this case. Indeed if we want to compute $g(x | S)$ as a probability, a natural approach would be to solve the following program ($\Omega$ is finite):

$$\text{find p which maximize } - \Sigma_{w_i \in \Omega} p(w_i) \log p(w_i) \text{ under the constraint } P(I_c) \geq \alpha \quad (21)$$

where $[\alpha, 1]$ is a numerical translation of "probable" appearing in S. The interval $I_c$ supposedly expresses the term "young". The solution of the above problem is simply, where |.| denotes cardinality:  $P(x|S) = \alpha / |I_c|$ if $x \in I_c$ ; $P(x|S) = (1 - \alpha) / (|\Omega| - |I_c|)$ if $x \notin I_c$ (22)

When $\alpha = 1$, we get a particular case of (20) where the a priori probability distribution is uniform. The above representation is good to <u>decide</u> what is Mary's age, not to represent our state of knowledge, when the a priori probability distribution is not available. In that case, statement S translates into $C(I_c) = \alpha$ where $C(I_c)$ is a lower probability degree, here a degree of belief in the sense of Shafer. We need some least information principle, which, in the scope of simply representing knowledge, would maximize the imprecision contained in the belief function Bel=C. A simple evaluation of the imprecision of the statement "Mary is young", where "young" translates into the interval $I_c$, is the cardinality of $I_c$. When $|I_c|=1$ then we know Mary's age, when $I_c = \Omega$ we have no information. More generally if Mary's age is known under the form of a random set $(\mathcal{F}, m)$, the imprecision is measured by

$$|(\mathcal{F}, m)| = \Sigma_{A \subseteq \Omega} m(A) \cdot |A| \quad (23)$$

which is an expected cardinality. Hence the problem of representing the meaning of a statement such as "Mary is probably young", where "probably" is viewed as specifying a lower bound on a probability value and "young" is viewed as a clearcut category approximated by a subset $I_c$, is that of solving the following program

$$\text{find m which maximizes } \Sigma_{A \subseteq \Omega} m(A) \cdot |A| \text{ under the constraint } \text{Bel}(I_c) = \Sigma_{A \subseteq I_c} m(A) = \alpha \quad (24)$$

This is the principle of minimum specificity [7]. The solution of (24) is easily found:

$$m(I_c | S) = \alpha \quad ; \quad m(\Omega | S) = 1 - \alpha \quad (25)$$

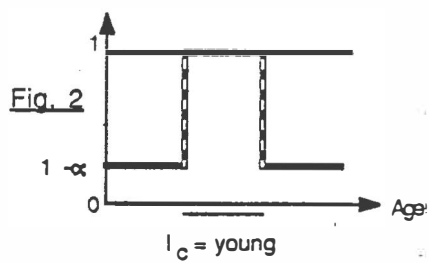

Fig. 2

$I_c$ = young

Note that this solution is a possibility measure whose distribution is $\pi(x|S) = 1$ if $x \in I_c$, $\pi(x|S) = 1 - \alpha$ otherwise (see Fig. 2). Moreover it encompasses the solution given by the maximum entropy principle as a particular case, namely

$$\forall A \subseteq \Omega \quad \text{Bel}(A|S) \leq P(A|S) \leq \text{Pl}(A|S) \quad (26)$$

i.e. the probability measure obtained by (22) is in the set of probability measures implicitly defined by Bel and Pl, viewed as upper and lower probabilities.

This approach to the representation of uncertain knowledge can be extended to the case when "young" is expressed, not as an interval, but as a fuzzy interval (e.g. [17]). Note that the maximum entropy principle can also accommodate fuzziness in the statement of constraints changing $P(I_c)$ into P(young) as defined by (14).

88

## Conclusion

This paper has tried to show that as far as knowledge representation is concerned, there should not be any dispute regarding the well-foundedness of possibility and fuzzy set theory, belief functions, versus probability theory. The latter is older and is presently far more developed than the two former. So, they are easy to criticize from the stand-point of probability theory. However the development of new models of uncertainty for knowledge representation seems to be an important issue, because of limitations of probability theory in terms of descriptive power. The new models such as possibility and evidence theories are not built <u>against</u> probability theory, but in the same spirit ; indeed most of the new uncertainty measures can be viewed as upper and lower probability measures. Hence probability theory itself is a basic tool for the construction of new models of uncertain knowledge. The present situation of probability theory is similar to the situation of classical logic in the mid-seventies. Classical logic has been given up by some researchers, when modelling common sense knowledge, but the idea of a logic has been preserved, and new logics (default logics for instance) have arisen. In the case of uncertainty models, it seems important to go beyond probability but consistently with probability theory itself. Of course the authors are aware of the bulk of work needed to bring possibility and evidence theories to the level of development of probability theory. In that sense many results reported in the literature about foundations and combination rules are certainly preliminary. But we strongly question dogmatic attitudes disputing alternative theories of uncertainty on behalf of rationality. Modelling uncertainty, and especially subjective uncertainty, cannot be but a compromise between the ideally optimal Bayesian theory and the limited precision and vagueness of the available often subjective knowledge.